\documentclass[journal]{IEEEtran}
\usepackage{amsmath,amssymb,amsfonts,color,subfigure}
\usepackage{graphicx}
\usepackage{setspace}
\usepackage{float}
\usepackage{epsfig,epstopdf}
\usepackage{verbatim}
\usepackage[section]{algorithm}
\usepackage{algorithmic}
\usepackage{array}
\usepackage{amsthm}
\usepackage{booktabs}
\usepackage{graphicx}
\usepackage{subfigure}
\usepackage{multirow}
\usepackage{threeparttable}
\usepackage{hyperref}
\usepackage{makecell}
\usepackage{fmtcount}
\usepackage{booktabs}

\usepackage{nicefrac}
\usepackage{xcolor}
\usepackage{xspace}
\usepackage{enumerate}
\usepackage{mathrsfs}



\newcolumntype{L}[1]{>{\raggedright\let\newline\\\arraybackslash\hspace{0pt}}m{#1}}
\newcolumntype{C}[1]{>{\centering\let\newline\\\arraybackslash\hspace{0pt}}m{#1}}
\newcolumntype{R}[1]{>{\raggedleft\let\newline\\\arraybackslash\hspace{0pt}}m{#1}}

\numberwithin{equation}{section}

\theoremstyle{remark}

\newcommand{\mat}[1]{\ensuremath{\mathbf{#1}}}

\newcommand{\eg}{e.g.}
\newcommand{\ie}{i.e.}

\newcommand{\cL}{\mathcal{L}}
\newcommand{\cD}{\mathcal{D}}

\newcommand{\R}{\mathbb{R}}

\newcommand{\scal}[2]{\left\langle #1,#2 \right\rangle}
\newcommand{\la}{\langle}
\newcommand{\ra}{\rangle}

\newcommand{\suml}[2]{\sum\limits_{#1}^{#2}}

\begin{document}
%
\title{Fast and Accurate Poisson Denoising with Optimized Nonlinear Diffusion}
%
%
%

\author{Wensen~Feng
        and~Yunjin~Chen
\thanks{
W.S. Feng is with School of Automation and Electrical Engineering, University of Science and Technology Beijing, Beijing,
100083, China. e-mail: (wensenfeng1986@gmail.com)

Y.J. Chen is with National Laboratory for Parallel and Distribution Processing
 National University of Defense Technology
 Deyalu 47, 410073 Changsha, Hunan Province, China
 e-mail: (chenyunjin$\_$nudt@hotmail.com)

\emph{Corresponding author: Yunjin Chen.}
}
}

\maketitle


\begin{abstract}
The degradation of the acquired signal by Poisson noise is a common problem for various imaging applications, such as medical imaging, night vision and microscopy.
Up to now, many state-of-the-art Poisson denoising techniques mainly
concentrate on achieving utmost performance, with little consideration for the
computation efficiency. Therefore, in this study we aim to propose an efficient Poisson denoising model with both
high computational efficiency and recovery quality.
To this end, we exploit the newly-developed trainable nonlinear reaction diffusion model which has proven
an extremely fast image restoration approach with performance surpassing recent state-of-the-arts.
We retrain the model parameters, including the linear
filters and influence functions by taking into account the Poisson noise statistics,
and end up with an optimized nonlinear diffusion model
specialized for Poisson denoising. The trained model provides strongly competitive results against state-of-the-art approaches,
meanwhile bearing the properties of simple structure and high efficiency.
Furthermore, our proposed model comes along with an additional advantage, that the diffusion process is well-suited for parallel computation on GPUs.
For images of size $512 \times 512$, our GPU implementation takes less than 0.1 seconds to produce state-of-the-art Poisson
denoising performance.
\end{abstract}

\section{Introduction}

\PARstart IMAGE degradation by Poisson noise is unavoidable in real applications such as
astronomy imaging, biomedical imaging, and microscopy, among many others \cite{Bertero2010, Rodrigues08, Berry05}.
Therefore, the Poisson noise removal is of crucial importance, especially
when data is to be submitted to further processing e.g. image segmentation
and recognition. Due to the physical mechanism, the strength of the Poisson noise depends on the image intensity and is therefore not additive, alluding to the fact that
Poisson denoising is generally quite different from the usual case of the additive noise.

Up to now, a host of Poisson denoising algorithms has been proposed in the literature, see \cite{INAnscombe7},
\cite{INAnscombe8}, \cite{Figueiredo}, \cite{INAnscombe10} and the references therein for a survey. Roughly speaking, major contributions consist of two classes: (1) with variance stabilizing transformation (VST)
and (2) without VST.

The approaches in the first class preprocess the input data by applying a nonlinear VST such as Anscombe \cite{Anscombe1,Anscombe2} or Fisz \cite{Fisz} which help remove the signal-dependency property of the Poisson noise. The noise characteristic of the transformed data can be approximately regarded as signal-independent additive Gaussian noise. Then, many well-studied Gaussian denoising algorithms can be employed to estimate a clean image for the transformed image. Finally, the estimate of the underlying noise-free image is obtained by applying an
inverse VST \cite{INAnscombe1, Lefkimmiatis, INAnscombe3, INAnscombe5} to the denoised transformed data. Using the well-known BM3D algorithm \cite{BM3D} for Gaussian noise removal, the resulting Poisson denoising algorithm leads to
state-of-the-art performance. However, the VST is accurate only when the measured pixels have relative high intensity.
That is to say, the recover error using the VST will dramatically increase for cases of low-counts \cite{Salmon2}, especially for extremely
low-count cases \eg, images with $\text{peak} = 0.1$.

In order to deal with the aforementioned deficiency of the VST operation, several authors \cite{Willettpoisson} \cite{Salmon2} \cite{Giryes} have investigated denoising strategies without VST, which rely directly on the statistics of the Poisson noise. In \cite{Salmon2}, J. Salmon $et$ $al.$ proposed a novel denoising algorithm in combination of
elements of dictionary learning and sparse patch-based representations, which relies directly on Poisson noise properties. It employs both an adaptation
of Principal Component Analysis (PCA) for Poisson
noise \cite{PoissonPCA1} and sparse Poisson intensity estimation methods \cite{SPIRAL-TAP} in a non-local framework. This direct approach achieves state-of-the-art results for images suffering from a high noise level. There are two versions involved in this method: the non-local PCA (NLPCA) and the non-local sparse PCA (NL-SPCA). Particularly, the NL-SPCA results in a better image restoration performance by integrating an $\ell_1$ regularization term to the minimized objective.

Similarly, to overcome the deficiency of VST, the data-fidelity term
originated from Poisson noise statistics is adopted in \cite{Figueiredo}, \cite{Le1}, \cite{Giryes}. Especially, the work in \cite{Giryes}
relies on the Poisson statistical model directly and uses a dictionary learning strategy with a sparse coding algorithm
that employs a boot-strapping based a stopping criterion. The reported
denoising performance of the method proposed in \cite{Giryes} is competitive with leading methods.

\subsection{Our Contribution}
While having a closer look at state-of-the-art Poisson denoising approaches, we find that such approaches
mainly concentrate on achieving utmost image restoration
quality, with little consideration on the computational efficiency.
A notable exception is the BM3D based algorithm incorporation with VST operation \cite{INAnscombe5}, which meanwhile offers high efficiency,
thanks to the highly engineered and well refined BM3D algorithm.

The goal of our work is to develop a simple but effective approach with both high computational efficiency and competitive denoising quality with state-of-the-art approaches. To this end, we employ the newly-developed trainable nonlinear reaction diffusion model \cite{chenCVPR15} which has several remarkable benefits. First, this model is merely a standard nonlinear diffusion model with trained filters and influence
functions, and therefore achieves very high levels of recovery quality surpassing
recent state-of-the-arts. Second, it needs only a small number of explicit steps, and hence is
extremely fast and highly computationally efficient. Furthermore, it is well suited
for parallel computation on GPUs. The employed model \cite{chenCVPR15} can also be interpreted as a recurrent convolutional neural networks (CNN) \cite{Gravesofflinehandwriting} with trainable activation functions. The standard CNN models typically fix the activation function and just train the filters. Differently, the employed model simultaneously optimizes the filters and the nonlinear activation functions, and therefore has a great potential to improve the performance of the standard CNN models.

In this paper, we start with an energy functional derived from the Poisson noise distribution,
 and derive a trainable nonlinear diffusion process specialized for the task of Poisson denoising. The model parameters in the diffusion process need to be trained by taking into account the Poisson noise statistics, including the linear filters and influence functions. Eventually, we reach a
nonlinear reaction diffusion based approach for Poisson denoising, which leads to state-of-the-art performance, meanwhile gains high computationally efficiency. Moreover, the straightforward direct gradient descent employed for Gaussian denoisng task \cite{chenCVPR15} is not applicable in our study. To solve this problem, we resort to the proximal gradient descent method
\cite{nesterov2004introductory}.
\subsection{Organization}
The remainder of the paper is organized as follows. Section II presents a general review of the statistics property of
Poisson noise, the trainable nonlinear reaction diffusion process and the proximal gradient method, which is required to derive the diffusion process for Poisson
denoising. In the subsequent section III, we propose the optimized nonlinear diffusion process for poisson noise reduction.
Subsequently, Section IV describes comprehensive experiment results for the proposed model.
The concluding remarks are drawn in the final Section V.
\section{Preliminaries}
\label{Preliminaries}
To make the paper self-contained, in this section we provide a brief review of Poisson noise, the trainable nonlinear diffusion process proposed in \cite{chenCVPR15} and the basic update rule of the proximal gradient algorithm \cite{nesterov2004introductory}.

\subsection{Poisson Noise}
Suppose that $f \in
{\mathbb{Z}_+^N}$ (represented as a column-stacked vector) denotes a Poisson noisy image and $u\in \mathbb{R}^N$ is the original true image of interest. Our task is to recover $u$ from $f$. Each observed pixel value $f_i$ in $f$ given $u_i$ in $u$ is assumed to be a Poisson
distributed independent random variable with mean and variance $u_i$, i.e.,
\begin{equation}
P\left( f_i| u_i\right) = \left\{
\begin{array}{ll}
{\frac{{u_i}^{f_i}}{f_i !}\mathrm{exp}\left( - u_i\right)}, & \mbox{$u_i>0$}\\
\delta_0\left(f_i \right), & \mbox{$u_i = 0,$}
\end{array} \right.
\label{poissondistri}
\end{equation}
where $u_i$ and $f_i$ are the $i$-th component in $u$ and $f$ respectively, and $\delta_0$ is
the Kronecker delta function. As is known, Poisson noise is signal dependent, due to the fact that the strength of the noise is proportional to the
signal intensity $u_i$. Therefore, the noise level in the image $u$ is generally defined
as the peak value (the maximal value) in $u$. This is reasonable since the effect of Poisson noise increases (\ie, the signal-to-noise ratio (SNR) decreases) as the intensity value
$u_i$ decreases.

Minimizing the negative log-likelihood $E = -\text{log}P(f|u)$ of \eqref{poissondistri} leads to the following data-fidelity term in the variational framework
\begin{equation}
\langle u-f \mathrm{log}u,1 \rangle,
\label{poissonfidelity}
\end{equation}
where $\langle,\rangle$ denotes the standard inner product.
This data-fidelity term (\ref{poissonfidelity}) is also known as the so-called Csisz{\'a}r I-divergence model \cite{csiszar1991least}, and
has been widely investigated in previous Poisson denoising algorithms, \eg, \cite{Salmon2,SPIRAL-TAP,Lingenfelter}.

\subsection{Trainable Nonlinear Reaction Diffusion}
A simple but effective framework for image restoration based on the concept of nonlinear
diffusion was recently proposed in our previous work \cite{chenCVPR15},
which extends conventional nonlinear reaction
diffusion models by several parameterized linear filters as well as several parameterized
influence functions. The proposed framework in \cite{chenCVPR15} is formulated as a time-dynamic nonlinear
reaction-diffusion model, having the following general form
\begin{equation}\label{diffusion}
\begin{cases}
u_0 = I_0 \\
u_{t+1} = u_t - \underbrace{\sum\limits_{i = 1}^{N_k}\bar k_i^t * \phi_i^t
(k_i^t * u_t)}_\text{diffusion term} - \underbrace{\psi(u_t, f)}_\text{reaction term}
,t = 0 \cdots T-1 \,,
\end{cases}
\end{equation}
where $*$ is the convolution operation, $T$ denotes the diffusion stages, 
$k_i$ are time varying convolution kernels 
($\bar k_i$ are obtained by rotating the kernel $k_i$ 180 degrees), 
$\phi_i^t$ are time varying
influence functions (not restricted to be of a certain kind), $\psi(u_t, f)$ is the reaction term, and $N_k$ is the number of filters. The reaction term can be typically chosen as
the derivative of a date term $\cD(u, f)$, i.e., $\psi(u) = \nabla_u \cD(u,f)$.
$I_0$ is the initial status of the diffusion process, and can be set as $f$. Note that the diffusion
behavior of each step in the model \eqref{diffusion} can be different.

The framework \eqref{diffusion} has wide applicability for various image restoration problems
by incorporating specific data terms. For example, it is easy to handle
classical image restoration problems, such as
Gaussian denoising, image deblurring, image super resolution and image
inpainting, by setting $\cD(u, f) =
\frac{\lambda}{2} \|A u - f\|_2^2$, $\psi(u) = \lambda A^\top(Au -
f)$,
where $f$ is the input degraded image, $A$ is the
associated linear operator\footnote{In the case of Gaussian denoising, $A$ is the
identity matrix; for image super resolution, $A$ is related to the down sampling operation and for image
deconvolution, $A$ corresponds to the linear blur kernel.},
and $\lambda$ is related to the strength of the reaction term.
For the problem of Poisson denoising exploited in this paper, the data term
should be chosen as follows according to \eqref{poissonfidelity}.
\begin{equation}
\cD(u, f) = \lambda \scal{u-f\text{log}u}{1}.
\label{datatermpoisson}
\end{equation}

At each time step $t$, the employed model \eqref{diffusion} can be interpreted as
performing one gradient descent step at $u^t$ with respect to a certain energy
functional given by
\begin{equation}\label{foemodel}
E(u, f) = \suml{i = 1}{N_k}\sum\limits_{p = 1}^{N} \rho_i^t((k_i^t
*u)_p) + \cD(u, f)\,,
\end{equation}
where the functions $\left\{\rho_i^t\right\}_{t=0}^{t=T-1}$ are the so-called penalty
functions. Note that $\rho'(z) = \phi(z)$ and
$k_i*u$ denotes 2D convolution of the image $u$
with the filter kernel $k_i$.
Since the parameters \{$k_i^t,
\rho_i^t$\} vary across the stages, \eqref{foemodel} is a dynamic energy
functional, which changes at each iteration.

\begin{figure*}[htb!]
\centering
\vspace{-0.5cm}
\hspace*{-0.9cm} {\includegraphics[width=1.1\linewidth]{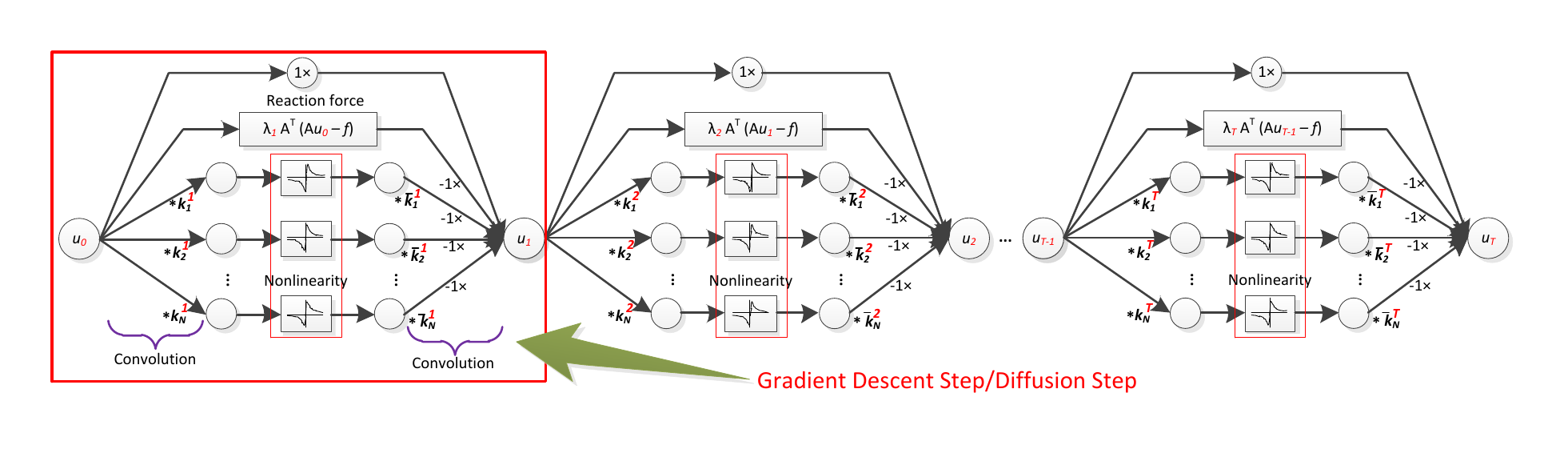}}
\vspace*{-1cm}
\caption{The architecture of the employed diffusion model with a reaction term, e.g., $\psi(u_{t-1}, f) = A^\top (A u_{t-1} - f)$.
It is represented as a feed-forward network. }\label{fig:feedforwardCNN}
\end{figure*}

The employed diffusion model is closely related to the
convolutional networks (CNs) employed for
image restoration problems \cite{CNNdenoising}. It can be treated as a convolutional network because each iteration (stage) of the diffusion process involves the convolution operation with a set of linear filters.
The architecture of the employed diffusion model is shown in Figure \ref{fig:feedforwardCNN}, and can be categorized into recurrent networks \cite{Gravesofflinehandwriting} with trainable activation functions. Besides, the special CNN architecture with trained activation functions has also been proposed in \cite{kaiminghe}. The work in \cite{kaiminghe} aims at ImageNet Classification and achieves surpassing Human-Level performance. However, the trainable activation function in \cite{kaiminghe} is parameterized by only one free variable, while in the employed diffusion model there are several free parameterized variables to learn.

The employed diffusion model is trained in a supervised manner,
namely we firstly prepare the input/output pairs
for certain image processing task, and then
exploit a loss minimization scheme to learn the model parameters $\Theta^t$ for each
stage $t$ of the diffusion process. The training dataset consists of
$S$ training samples $\{f^s,u_{gt}^s\}_{s=1}^S$, where
$f^s$ is a degraded input and $u_{gt}^s$ is the corresponding ground truth clean image.
The model parameters $\Theta^t$ of each stage include the parameters of (1) the reaction force weight $\lambda$,
(2) linear filters and (3) influence functions, i.e., $\Theta^t = \{\lambda^t, \phi_i^t, k_i^t\}$. The training task is formulated as the following
optimization problem
\begin{equation}\label{learning}
\hspace{-0.75cm}\begin{cases}
\min\limits_{\Theta}\cL(\Theta) = \sum\limits_{s = 1}^{S}\ell\left( u_T^s , u_{gt}^s \right) =\sum\limits_{s = 1}^{S}\frac 1 2
\|u_T^s - u_{gt}^s\|^2_2\\
\text{s.t.}
\begin{cases}
u_0^s = f^s \\
u_{t+1}^s = u_t^s - {\sum\limits_{i = 1}^{N_k}\bar k_i^t * \phi_i^t
(k_i^t * u_t^s)} - \psi(u_t^s, f^s), t = 0 \cdots T-1\,,
\end{cases}
\end{cases}
\end{equation}
where $\Theta = \{\Theta^t\}_{t=0}^{t=T-1}$. The training problem can be solved via gradient
based algorithms, e.g., commonly used L-BFGS algorithm \cite{lbfgs}.
The gradients of the loss function with respect to $\Theta_t$ are computed
using the standard back-propagation technique widely used in
the neural networks learning \cite{lecun1998gradient}. There are two training strategies to learn the diffusion processes: 1) the greedy training strategy to learn the diffusion process stage-by-stage; and 2) the joint training strategy to joint train all the stages simultaneously. Generally speaking, the joint training strategy performs better \cite{chenCVPR15}, and the greedy training strategy is often employed to provide a good initialization for the joint training. For simplicity, we just consider the joint training scheme to
train a diffusion process by simultaneously tuning
the parameters in all stages. The associated gradient $\frac {\partial \ell(u_T, u_{gt})}{\partial \Theta_t}$ is presented as follows,
\begin{equation}
\frac {\partial \ell(u_T, u_{gt})}{\partial \Theta_t} =
\frac {\partial u_t}{\partial \Theta_t} \cdot \frac {\partial u_{t+1}}{\partial u_{t}} \cdots
\frac {\partial \ell(u_T, u_{gt})}{\partial u_T} \,.
\label{iterstep}
\end{equation}
One can refer to \cite{chenCVPR15} and its supplementary materials for the detailed calculation process of $\frac {\partial u_t}{\partial \Theta_t}$ and $\frac {\partial u_{t+1}}{\partial u_{t}}$.
\subsection{The Proximal Gradient Method}
In order to derive the diffusion process for Poisson denoising, we
start with the energy functional \eqref{foemodel} by incorporating the data term
\eqref{datatermpoisson}. However, for this problem, the straightforward
direct gradient descent $1-\frac{f}{u}$ is not applicable in practice,
because (1) it has an evident problem of numerical instability at the points
with $u$ very close to zero; (2) this update rule can not guarantee that
the output image after one diffusion step is positive. Negative values of
$u$ will violate the constraint of the data term \eqref{datatermpoisson}.

As a consequence, we resort to the proximal gradient descent method
\cite{nesterov2004introductory}, which can avoid the formula $1-\frac{f}{u}$, and
thus solve the above two problems.
The proximal gradient method is applicable to solve an optimization problem which is composed of
a smooth function $F$ and a convex (possibly non-smooth) function $G$:
\begin{equation}
\label{pg}
\arg\min\limits_{u}F(u) + G(u).
\end{equation}
It is based on a forward-backward splitting scheme. The basic update
rule to solve \eqref{pg} is given as
\begin{equation}\label{pgupdate}
u^{n+1}=\left( \mat{I}+\tau \partial G \right)^{-1}\left( u^n-\tau \nabla F\left( u^n \right) \right),
\end{equation}
where $\tau$ denotes the step size parameter, and $u^n-\tau \nabla F\left( u^n \right)$ is the forward gradient descent step.
The term $\left( \mat{I}+\tau \partial G \right)^{-1}$ denotes the
standard proximal mapping \cite{nesterov2004introductory}, and is also the backward step.
The proximal mapping $\left( \mat{I}+\tau \partial G \right)^{-1}(\tilde u)$ with respect to $G$ is given as the following
minimization problem
\begin{equation}\label{subproblemG}
\left( \mat{I} + \tau \partial G \right)^{-1}(\tilde{u}) = \arg\min\limits_{u} \frac{\|u - \tilde
{u}\|^2_2}{2} + \tau G(u)\,.
\end{equation}

\section{Optimized Nonlinear Diffusion Process for Poisson Noise Reduction}
\label{section3}
\subsection{Proposed Diffusion Process for Poisson Denoising}
In this section, we propose the optimized nonlinear reaction diffusion process for the task of Poisson denoising.
First of all, it should be noted that the diffusion process can be interpreted as one gradient decent step of the energy functional \eqref{foemodel}.
Therefore, we start from the following variational model by
combing the data-fidelity term \eqref{datatermpoisson},
\begin{equation}\label{straightforward}
\arg\min\limits_{u > 0}E(x) =
\suml{i=1}{N_k} \sum\limits_{p = 1}^{N} \rho_i((k_i
*u)_p) + \lambda \langle u-f \mathrm{log}u,1 \rangle \,.
\end{equation}
As mentioned earlier, the straightforward direct gradient descent is
not applicable for this problem. Therefore, we resort to the proximal gradient
descent method to derive the corresponding diffusion process.
Casting the problem \eqref{straightforward} in the form of \eqref{pg}, we have $F(u)=\suml{i=1}{N_k} \sum\limits_{p = 1}^{N} \rho_i((k_i
*u)_p)$ and $G(u)=\lambda \langle u-f \mathrm{log}u,1 \rangle$. It is easy to check that
\[
\nabla F(u)= \suml{i=1}{N_k} K_i^\top \phi_i(K_iu),
\]
where $K_i \in \R^{N \times N}$ is a highly sparse matrix, implemented as 2D convolution of the image $u$
with the filter kernel $k_i$, i.e., $K_i u \Leftrightarrow k_i*u$, $\phi(K_i u)
= \left( \phi(K_i u)_1, \cdots, \phi(K_i u)_N \right)^\top
\in \R^N$, and $\rho'(z) = \phi(z)$. 
Note that in the case of symmetric boundary condition exploited in our work, 
$K_i^\top$ can be interpreted as the convolution kernel 
$\bar{k}_i$ only in the central region. 
As stated in \cite{chenCVPR15}, in order to simplify the diffusion model, 
it is better to revise the above formulation as explicit convolutions. 
Therefore, the gradient of $F$ is revised as
\[
\nabla F(u) \doteq \suml{i=1}{N_k} \bar{k}_i * \phi_i(k_i * u) \,.
\]
Recall that $\bar k_i$ is a rotated version of 180 degrees of kernel $k_i$.

The proximal mapping with respect to $G$ is given as the following minimization problem
\begin{equation}\label{subproblemGpoisson}
\left( \mat{I} + \tau \partial G \right)^{-1}(\tilde{u}) = \arg\min\limits_{u} \frac{\|u - \tilde
{u}\|^2_2}{2} + \tau \lambda \langle u-f \mathrm{log}u,1 \rangle \,.
\end{equation}
The solution of \eqref{subproblemGpoisson} is given by the following point-wise operation
\begin{equation}\label{subproblemGIdiv}
\hat u = \left( \mat{I} + \tau \partial G \right)^{-1}(\tilde{u})  = \frac{ \tilde{u} - \tau \lambda
+\sqrt{\left( \tilde{u} -\tau \lambda \right)^2+4\tau \lambda f}}{2} \,.
\end{equation}
Note that $\hat u$ is always positive if $f > 0$, i.e, this update rule can
guarantee $\hat u > 0$ in diffusion steps.

As a consequence, the diffusion process for Poisson denoising using the proximal gradient method can be formulated as
\begin{equation}\label{diffusionprocessfinal}
u_{t+1}  = \frac{ \tilde{u}_{t+1} - \lambda^{t+1}
+\sqrt{\left( \tilde{u}_{t+1} - \lambda^{t+1} \right)^2+4 \lambda^{t+1} f}}{2} \,,
\end{equation}
where $\tilde{u}_{t+1} =
u_t - \suml{i=1}{N_k}\bar{k}_i^{t+1} * \phi_i^{t+1}(k_i^{t+1} * u_{t})$, and we set
the step size $\tau = 1$. 

\subsection{Computing The Gradients for Training}
In this subsection, we present the joint training strategy for poisson denoising.

First of all, the diffusion equation for Poisson denoising is as presented in \eqref{diffusionprocessfinal}, from which we can compute the gradients of the loss function w.r.t the training parameters $\Theta_t = \left\{ \lambda^t,\phi_i^t, k_i^t \right\}$. According to \eqref{iterstep}, we should compute three parts of $\frac {\partial \ell(u_T, u_{gt})}{\partial \Theta_t}$, i.e., $\frac {\partial u_{t+1}}{\partial u_{t}}$, $\frac {\partial u_t}{\partial \Theta_t}$ and $\frac {\partial \ell(u_T, u_{gt})}{\partial u_T}$.

First of all, the gradients $\frac {\partial \ell(u_T, u_{gt})}{\partial u_T}$ is
easy to calculate according to the training loss function. For example, in the case of
quadratic training cost function, $\frac {\partial \ell(u_T, u_{gt})}{\partial u_T}$ is given as
\[
\frac {\partial \ell(u_T, u_{gt})}{\partial u_T} = u_T - u_{gt}\,.
\]

According to the chain rule,
$\frac{\partial u_{t+1}}{\partial u_t}$ is computed as follows,
\begin{equation}\label{derv1}
\frac{\partial u_{t+1}}{\partial u_t} =\frac{\partial \tilde{u}_{t+1}}{\partial u_t} \cdot \frac{\partial u_{t+1}}{\partial \tilde{u}_{t+1}}.
\end{equation}
Starting from the update rule \eqref{diffusionprocessfinal},
it is easy to check that $\frac{\partial \tilde{u}_{t+1}}{\partial u_t}$
is given as
\begin{equation}\label{derv1-1}
\begin{array}{l}
\frac{\partial \tilde{u}_{t+1}}{\partial u_t} = \mat{I} -
\suml{i=1}{N_k} {K_i^{t+1}}^\top \cdot \Lambda_i \cdot {\left({\bar{K}_i}^{t+1}\right)}^\top.
\end{array}
\end{equation}
where $\Lambda_i$ is a diagonal matrix
$\Lambda_i = \text{diag}({\phi_i^t}'(x_1), \cdots, {\phi_i^t}'(x_N))$ (${\phi_i^t}'$ is the first order derivative of function ${\phi_i^t}$), with
$x = {k_i^{t+1}} * u_t$. Here, $\{ x_i \}_{i=1}^{i=N}$ denote the element of $x$ which is represented as a column-stacked vector.
Note that in practice, we do not need to explicitly construct the
matrices $K_i$ and ${\bar{K}_i}$.
As shown in \cite{chenCVPR15}, $K_i^\top$ and
${\bar{K}_i}^\top$ can be computed by the convolution operation
with the kernel $k_i$ and $\bar{k}_i$, respectively with careful boundary handling.

Then the part $\frac{\partial u_{t+1}}{\partial \tilde{u}_{t+1}}$ can be computed according to \eqref{diffusionprocessfinal} and is formulated as
\begin{equation}\label{derv2}
\frac{\partial u_{t+1}}{\partial \tilde{u}_{t+1}} =
\text{diag}(y_1, \cdots, y_N)\,,
\end{equation}
where $\{ y_i \}_{i=1}^{i=N}$ denote the elements of
\[
y = \frac{1}{2} \left[ 1+\frac{\tilde{u}_{t+1}-\lambda^{t+1}}
{\sqrt{\left( \tilde{u}_{t+1}-\lambda^{t+1} \right)^2+4\lambda^{t+1} f}} \right]\,.
\]
Now, the $\frac{\partial u_{t+1}}{\partial u_t}$ is obtained by combing \eqref{derv1-1} and \eqref{derv2}.

Concerning the gradients $\frac{\partial u_t}{\partial \Theta_t}$,
as $\Theta_t$ involves $\left\{ \lambda^t,\phi_i^t, k_i^t \right\}$, we can derive them respectively. It is worthy noting that the gradients of $u_t$ w.r.t $\left\{\phi_i^t, k_i^t \right\}$ are only associated with
$\tilde{u}_t=u_{t-1} - \suml{i=1}{N_k}\bar{k}_i^t * \phi_i^t(k_i^t * u_{t-1})$. Therefore, the gradient of $u_t$ w.r.t $\phi_i^t$ and $k_i^t$ is computed via
\[
\frac{\partial u_t}{\partial \phi_i^t} = \frac{\partial \tilde{u}_t}{\partial \phi_i^t} \cdot  \frac{\partial u_t}{\partial \tilde{u}_t},
\]
and
\[
\frac{\partial u_t}{\partial k_i^t} =  \frac{\partial \tilde{u}_t}{\partial k_i^t} \cdot \frac{\partial u_t}{\partial \tilde{u}_t},
\]
where the derivations of $\frac{\partial \tilde{u}_t}{\partial \phi_i^t}$ and $\frac{\partial \tilde{u}_t}{\partial k_i^t}$ have been provided in \cite{chenCVPR15}.
The gradients of $\frac{\partial u_t}{\partial \tilde{u}_t}$ are calculated
similar to \eqref{derv2}.
The gradient of $u_t$ w.r.t $\lambda^t$ is computed as
\begin{equation}\label{derv4}
\frac{\partial u_t}{\partial \lambda^t} =
(z_1, \cdots, z_N)\,,
\end{equation}
where $\{ z_i \}_{i=1}^{i=N}$ denote the elements of
\[
z = \frac{1}{2} \left[ -1+\frac{\left(\lambda^t-\tilde{u}_t\right)+2f}{\sqrt{\left( \tilde{u}_t-\lambda^t \right)^2+4\lambda^t f}} \right]\,.
\]
Note that $\frac{\partial u_t}{\partial \lambda^t}$ is written as
a column vector.

In practice, in order to ensure the value of $\lambda^t$ positive during
the training phase, we set $\lambda=e^{\beta}$.
As a consequence, in the programming we employ the gradient $\frac{\partial u_t}{\partial \beta^t}$ instead of $\frac{\partial u_t}{\partial \lambda^t}$. The gradient $\frac{\partial u_t}{\partial \beta^t}$ can be explicitly formulated as
\begin{equation}\label{derv5}
\frac{\partial u_t}{\partial \beta^t} = \frac{\lambda^t}{2} \left[ -1+\frac{\left(\lambda^t-\tilde{u}_t\right)+2f}{\sqrt{\left( \tilde{u}_t-\lambda^t \right)^2+4\lambda^t f}} \right].
\end{equation}

\section{Experiments}
In this section, we considered the fully Trained Reaction Diffusion models for Poisson Denoising (TRDPD). The corresponding nonlinear diffusion process of stage $T$ with filters of size $m\times m$ is expressed as $\mathrm{TRDPD}_{m \times m}^T$ whose number of filters is $m^2-1$ in each stage, if not specified.

To generate the training data for our denoising experiments, we cropped a 180$\times$180 pixel region
from each image of the Berkeley segmentation
dataset \cite{MartinFTM01}, resulting in a total of 400 training samples of size 180 $\times$ 180. Of course, we also employ different amounts of training samples (e.g., 50-600 in Fig.~\ref{examples}) to observe the denoising performance comparison.

After training the models, we evaluate them on 68 test images originally introduced by
\cite{RothFOE}, which have since become a reference set for image
denoising. To provide a comprehensive comparison, the test peak values are distributed between 1 to 40.

Moreover, the proposed algorithm is compared with two representative state-of-the-art methods:
the NLSPCA \cite{Salmon2} and BM3D-based method with the exact unbiased inverse Anscombe
\cite{INAnscombe5}, both with and without binning technique.
The corresponding codes are downloaded from the author's homepage, and we use them as is.
For the binning technique, we closely follow \cite{Salmon2} and
use a $3 \times 3$ ones kernel to increase the peak value to be 9 times higher, and
a bilinear interpolation for the upscaling of the low-resolution
recovered image.
Two commonly used quality measures are taken to evaluate the Poisson denoising performance, \ie, PSNR and
the structural similarity index (SSIM) \cite{ssim}. Note that the PSNR values in the following three subsections are evaluated by averaging denoised results of 68 test images. For simplicity the test peak value is set as 40 in the following three contrastive analysises, without loss of generality.
\subsection{Influence of Number of Training Samples}
In this subsection, we evaluate the test performance of trained models using different amounts of training samples for $\mathrm{TRDPD}_{5 \times 5}^5$.

\begin{figure}[htbp]
\centering
\includegraphics[width=0.4\textwidth]{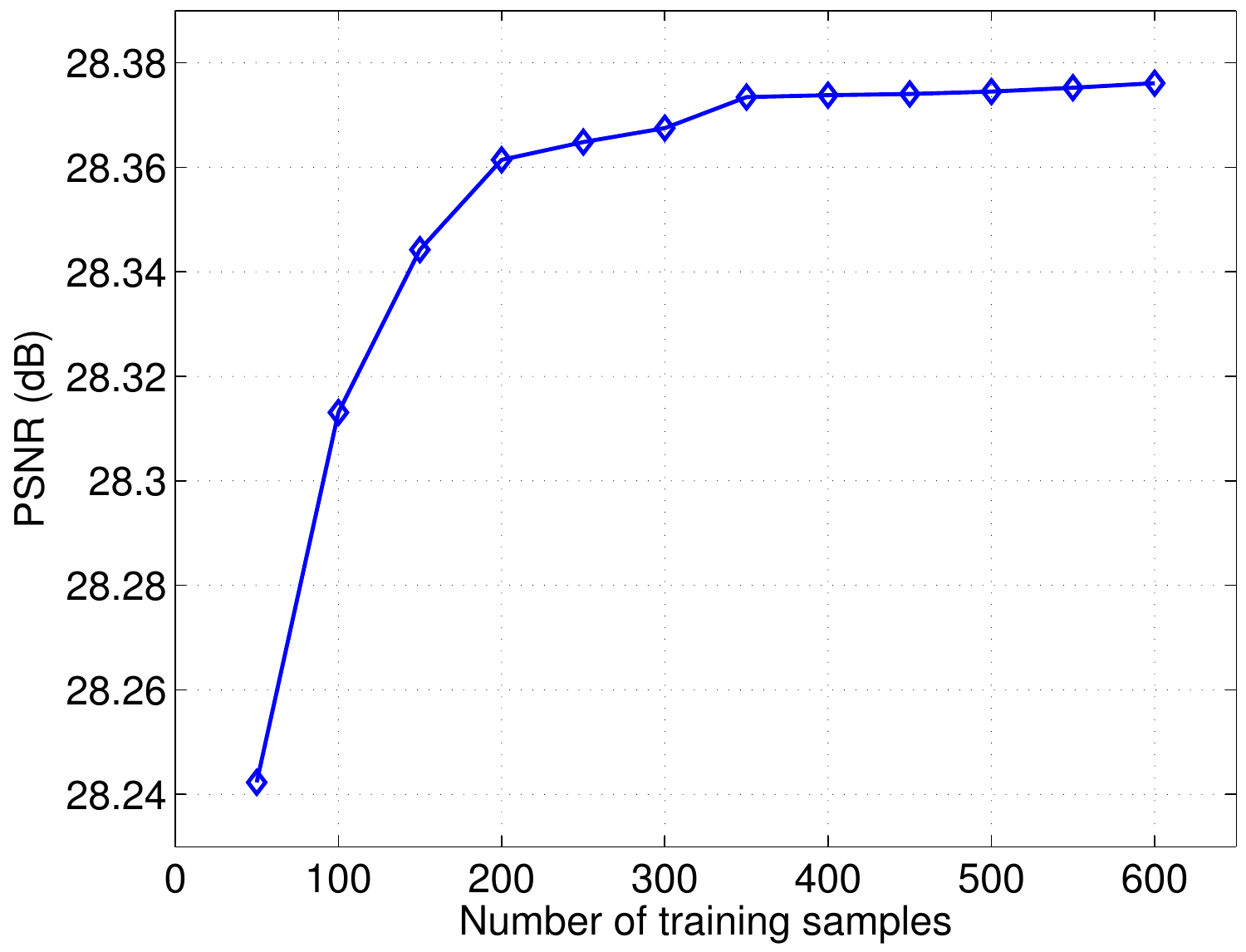}
\caption{Influence of the number of training examples.}
\label{examples}
\end{figure}

The results are summarized in Fig.~\ref{examples}, from which one can see that 350-400 images are typically enough to provide reliable performance. It is also worthy noting that too small training set will result in over-fitting which leads to inferior PSNR value.

\subsection{Influence of filter size}
\begin{figure}[htbp]
\centering
\includegraphics[width=0.4\textwidth]{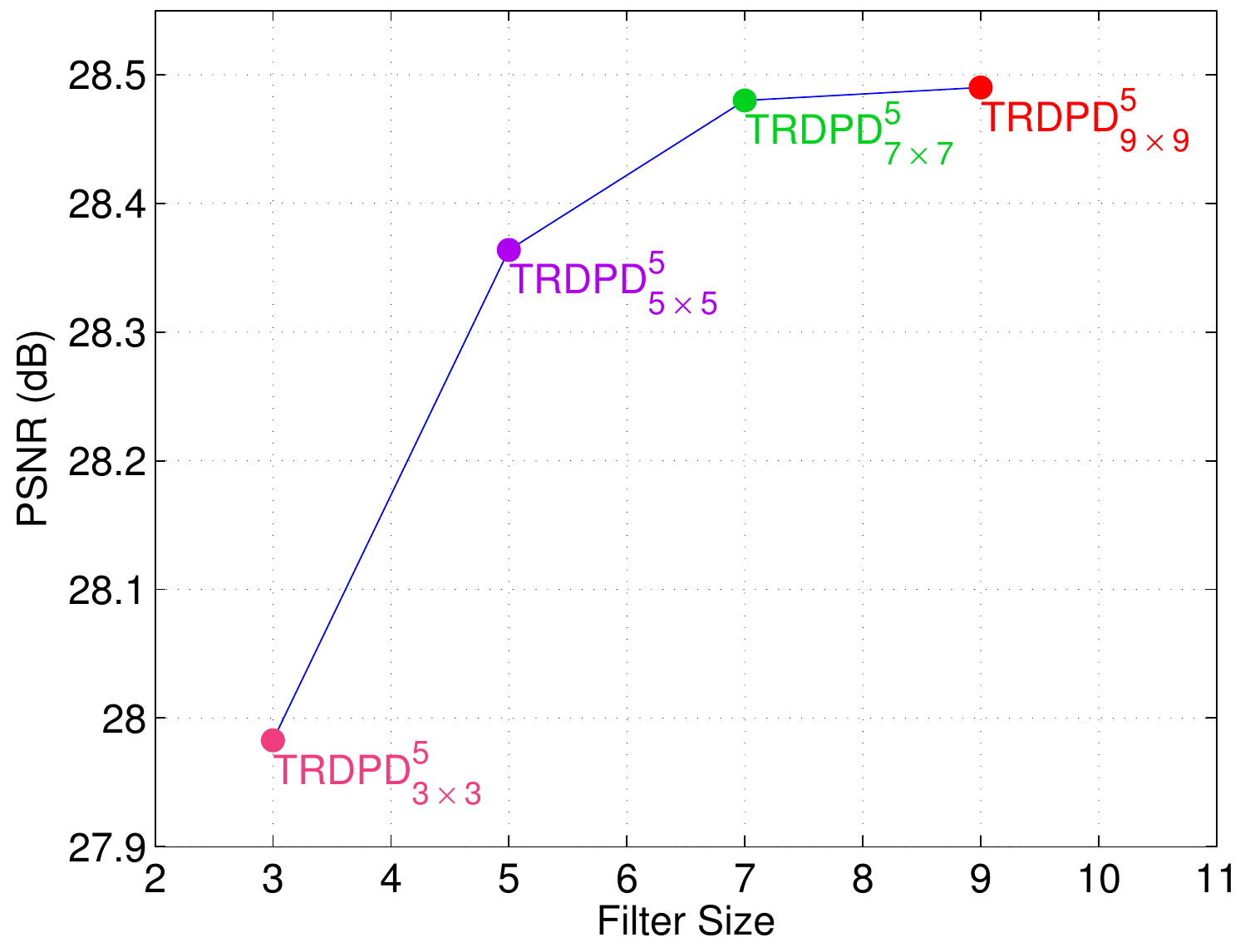}
\caption{Influence of the filter size.}
\label{filtersize}
\end{figure}

We also investigate the influence of the filter size on the denoising performance in Fig.~\ref{filtersize}. The diffusion stages are set as 5, and 400 images are used for training.

One can see that increasing
the filter size from $3\times 3$ to $5 \times 5$ brings a significant improvement. However, a much smaller PSNR gain ($\leq 0.01$) is achieved if we increase the filter size from $7\times 7$ to $9 \times 9$. By evaluating the training time and the performance improvement, we prefer the $\mathrm{TRDPD}_{7 \times 7}^T$ model as it provides
the best trade-off between performance and training time.

\subsection{Influence of Diffusion Stages}
In this study, any number of diffusion stages can be exploited in our model. But in practice, the trade-off between run time and accuracy should be considered. Therefore, we need to study the influence of the number of diffusion stages on the denoising performance. $\mathrm{TRDPD}_{5 \times 5}^T$ and 400 images are used for training.

\begin{figure}[htbp]
\centering
\includegraphics[width=0.4\textwidth]{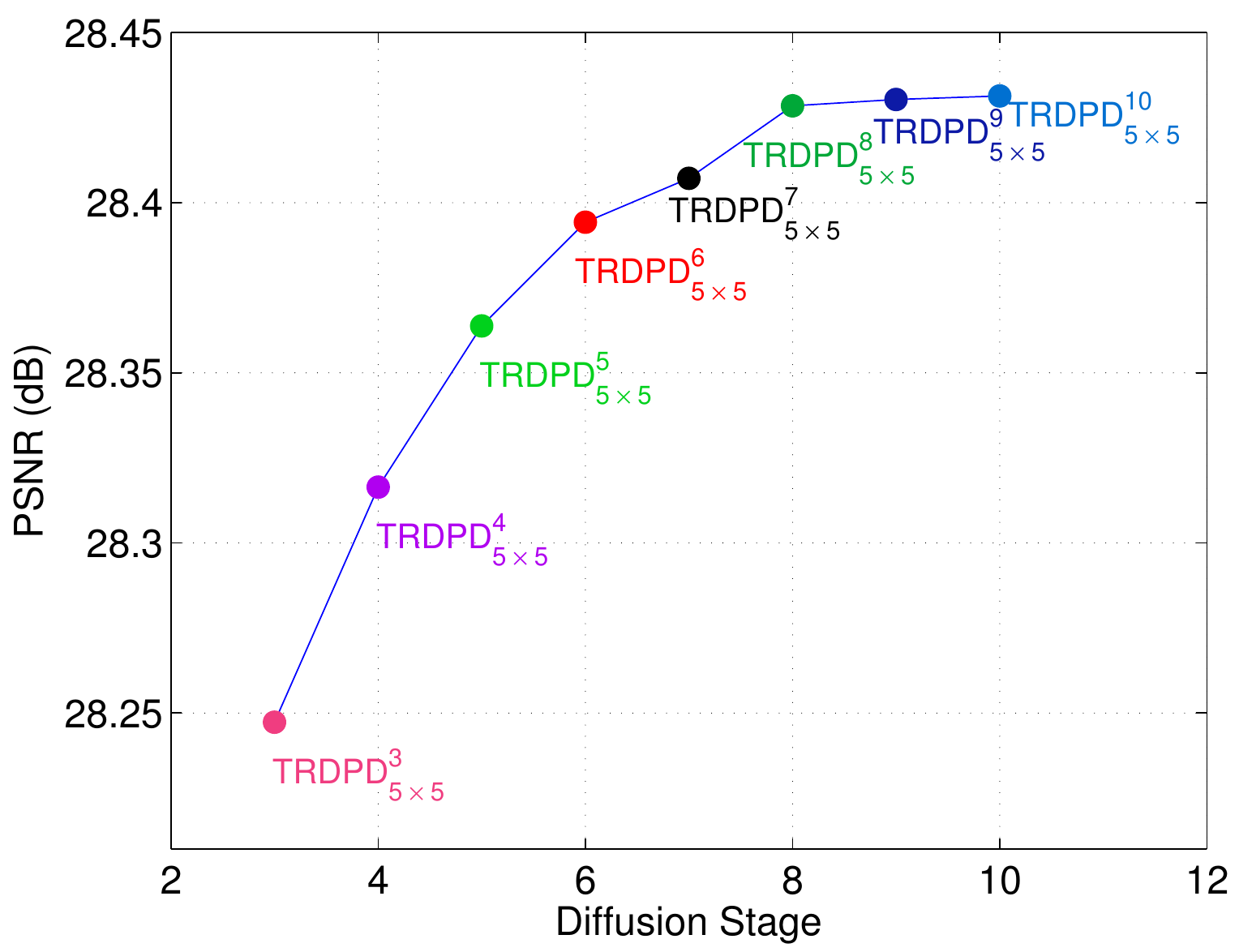}
\caption{Influence of the number of diffusion stages.}
\label{stages}
\end{figure}

As shown in Fig.~\ref{stages}, the performance improvement becomes insignificant ($\leq0.05$) when the diffusion stages $\geq 8$. In order to save the training time, we choose $\mathrm{Diffusion  \; Stage} = 8$ in the following experiments as it provides
the best trade-off between performance and computation time.

\begin{table*}[htbp]
\centering
\begin{tabular}{|c |c |c |c |c |c |c |}
\hline
Method & Peak=1 &Peak=2 &Peak=4 &Peak=8 &Peak=20 &Peak=40\\
\hline
\hline
NLSPCA & 20.90/0.491 &21.60/0.517&22.09/0.535&22.38/0.545&22.54/0.549&22.56/0.550\\
\hline
NLSPCAbin & 19.89/0.466 &19.95/0.467&19.95/0.467&19.91/0.466&19.72/0.463&19.37/0.459\\
\hline
BM3D & 21.01/0.504 &22.21/0.544&23.54/0.604&24.84/0.665&26.67/0.745&28.20/0.801\\
\hline
BM3Dbin & 21.39/0.515 &22.14/0.542&22.87/0.571&23.53/0.602&24.25/0.642&24.67/0.667\\
\hline
$\mathrm{TRDPD}_{5 \times 5}^8$& 21.49/0.512 &22.54/0.557&23.70/0.610&24.96/0.670&26.88/0.754&28.42/0.809\\
\hline
$\mathrm{TRDPD}_{7 \times 7}^8$& \textbf{21.60}/\textbf{0.518} &\textbf{22.62}/\textbf{0.560}&\textbf{23.84}/\textbf{0.618}&\textbf{25.14}/\textbf{0.680}&\textbf{26.98}/\textbf{0.759}&\textbf{28.50}/\textbf{0.812}\\
\hline
\end{tabular}
\caption{Comparison of the performance of the test algorithms in terms of PSNR and SSIM. Best results are marked.}
\label{resultshow}
\end{table*}
\subsection{Experimental Results}
By analyzing the above three subsections, we decide to employ $\mathrm{TRDPD}_{7 \times 7}^8$ model and 400 images for training. Meanwhile, we also test the $\mathrm{TRDPD}_{5 \times 5}^8$ model for comparison. Note that, the diffusion model needs to be trained respectively for different noise levels.

\begin{figure*}[htbp]
\centering
\subfigure[Image 1]{
\centering
\includegraphics[width=0.22\textwidth]{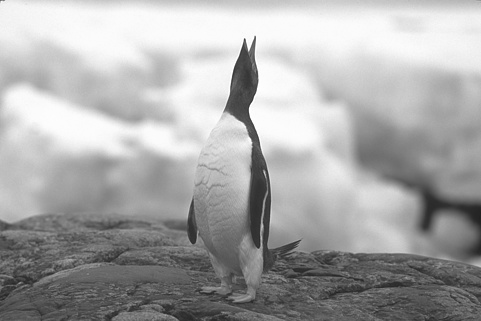}
}
\subfigure[Noisy image. Peak=1]{
\centering
\includegraphics[width=0.22\textwidth]{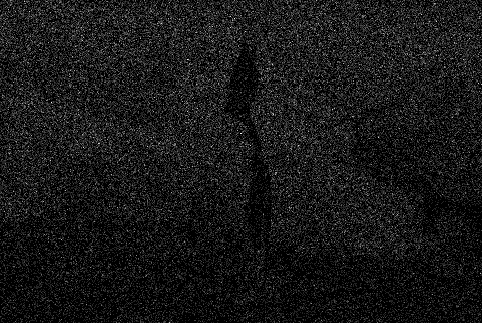}
}
\subfigure[NLSPCA (24.09/0.682)]{
\centering
\includegraphics[width=0.22\textwidth]{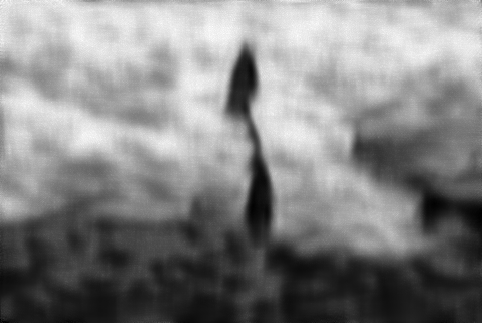}
}
\subfigure[NLSPCAbin (22.77/0.719)]{
\centering
\includegraphics[width=0.22\textwidth]{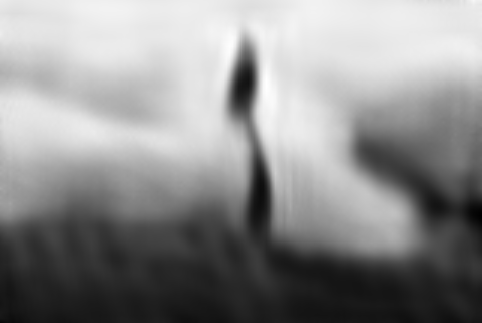}
}
\\
\subfigure[BM3D (22.92/0.636)]{
\centering
\includegraphics[width=0.22\textwidth]{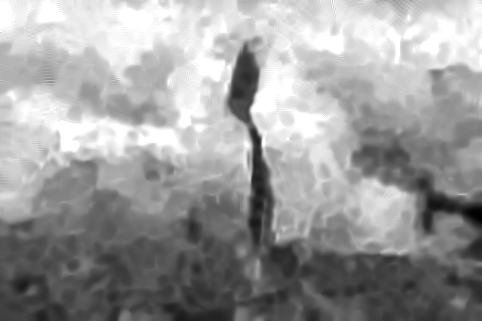}
}
\subfigure[BM3Dbin (24.57/\textbf{0.730})]{
\centering
\includegraphics[width=0.22\textwidth]{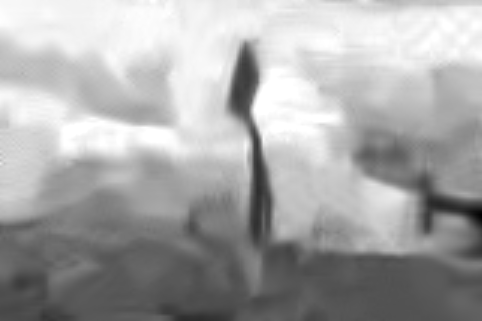}
}
\subfigure[$\mathrm{TRDPD}_{5 \times 5}^8$ (24.65/0.701)]{
\centering
\includegraphics[width=0.22\textwidth]{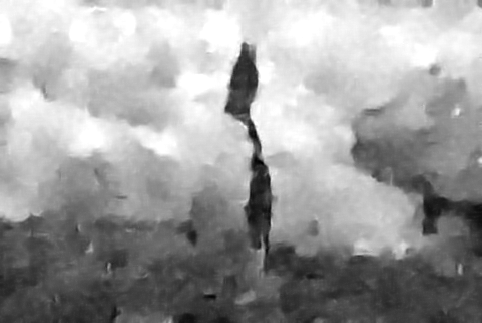}
}
\subfigure[$\mathrm{TRDPD}_{7 \times 7}^8$ (\textbf{24.94}/0.725)]{
\centering
\includegraphics[width=0.22\textwidth]{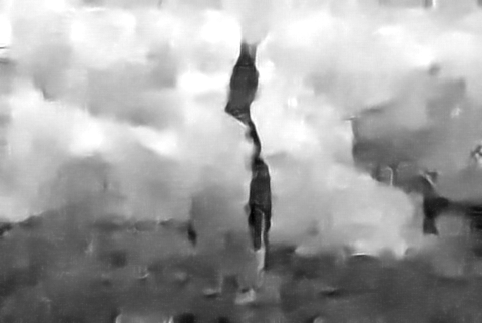}
}
\caption{Denoising of Image 1 with peak=1. The results are reported by PSNR/SSIM index. Best results are marked.}
\label{peak1}
\end{figure*}

\begin{figure*}[htbp]
\centering
\subfigure[Image 2]{
\centering
\includegraphics[width=0.22\textwidth]{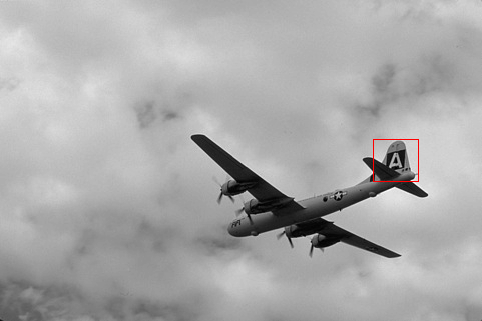}
}
\subfigure[Noisy image. Peak=1]{
\centering
\includegraphics[width=0.22\textwidth]{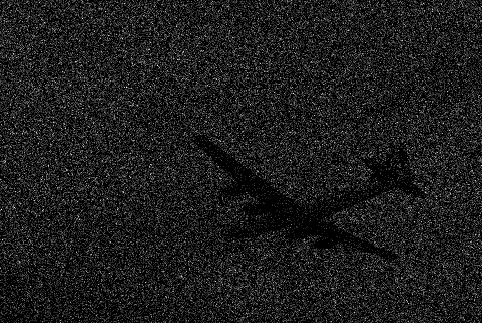}
}
\subfigure[NLSPCA (24.76/0.858)]{
\centering
\includegraphics[width=0.22\textwidth]{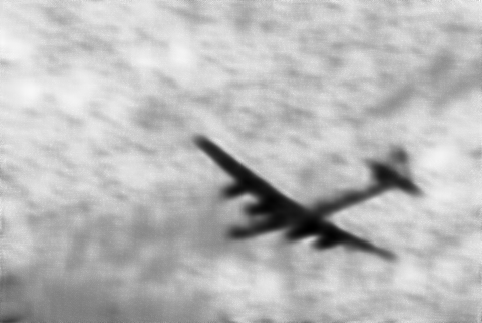}
}
\subfigure[NLSPCAbin (23.25/0.908)]{
\centering
\includegraphics[width=0.22\textwidth]{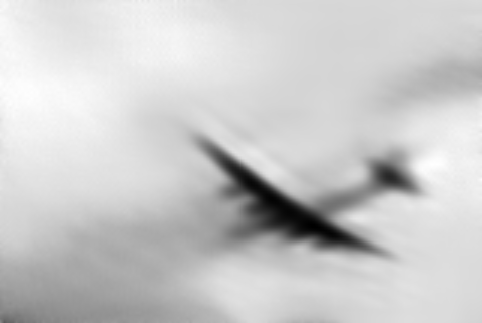}
}
\\
\subfigure[BM3D (24.23/0.881)]{
\centering
\includegraphics[width=0.22\textwidth]{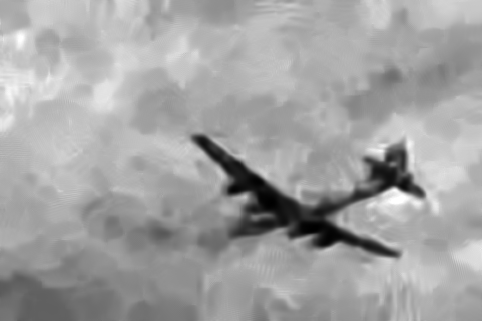}
}
\subfigure[BM3Dbin (25.66/\textbf{0.913})]{
\centering
\includegraphics[width=0.22\textwidth]{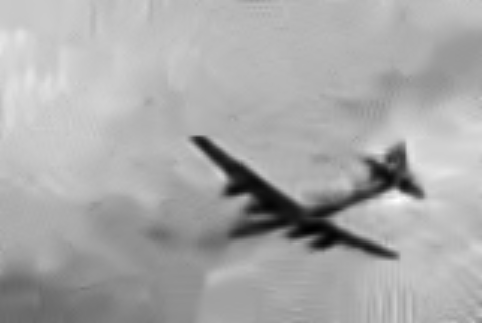}
}
\subfigure[$\mathrm{TRDPD}_{5 \times 5}^8$ (25.53/0.876)]{
\centering
\includegraphics[width=0.22\textwidth]{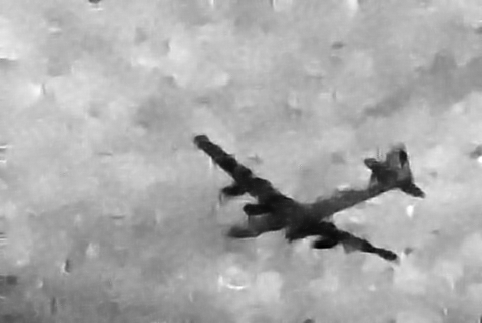}
}
\subfigure[$\mathrm{TRDPD}_{7 \times 7}^8$ (\textbf{26.13}/0.900)]{
\centering
\includegraphics[width=0.22\textwidth]{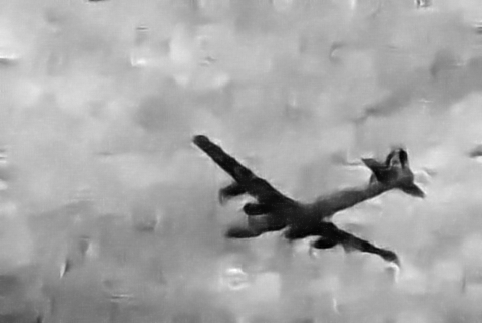}
}
\caption{Denoising of Image 2 with peak=1. The results are reported by PSNR/SSIM index. Best results are marked.}
\label{peak1-2}
\end{figure*}

Examining the recovery images in Fig.~\ref{peak1} and Fig.~\ref{peak1-2}, we see that in comparison with the other methods, the proposed algorithm is more accurate at capturing details, especially the recovery results within the red rectangle.

As shown in Fig.~\ref{peak1}(e)-Fig.~\ref{peak1-2}(e) the nonlocal technique BM3D is affected by the structured signal-like artifacts that appear in homogeneous areas of the image. This phenomenon is originated from the selection process of similar image patches in the BM3D denoising scheme. The selection process is easily influenced by the noise itself, especially in flat areas of the image, which can be dangerously self-referential. Therefore, the BM3D-based method without binning brings in the typical structured artifacts since most parts of images Image1 and Image2 are homogeneous, as shown in Fig.~\ref{peak1}(e) and Fig.~\ref{peak1-2}(e). However, the binning technique yields noisy images with lower noise level, thereby the denoised results in this case is less disturbed by the structured signal-like artifacts. Meanwhile, we find that our method introduces block-type artifacts if the peak values are relatively low, e.g., $\mathrm{peak}=1$ in Fig.~\ref{peak1}(g). The main reason is that our method is a local model, which becomes less effective to infer the underlying structure solely from the local neighborhoods, if the input image is too noisy.

It is also worthy noting that the SSIM values obtained by NLSPCA and BM3D using the binning technique in Fig.~\ref{peak1} and Fig.~\ref{peak1-2} are slight better than TRDPD. The binning operation results in a smaller Poisson image with lower resolution but higher counts per pixel. Thereby, in the extreme noise level case, the binning technique yields a significant performance
increase for some images. However, the adoption of the binning technique leads to resolution reduction which will weaken or even eliminate many image details. Therefore, for the images whose most parts are homogeneous (e.g., Image1 and Image2), the recover quality will be enhanced using the binning operation because images of this kind have few easily-missed details. Overall speaking, the performance of TRDPD in terms of PSNR/SSIM is better than the other methods for peak=1, as shown in Table~\ref{resultshow}. This indicates that for most images our method is more powerful in the recover quality and geometry feature preservation.

\begin{figure*}[htbp]
\centering
\subfigure[Image 3]{
\centering
\includegraphics[width=0.22\textwidth]{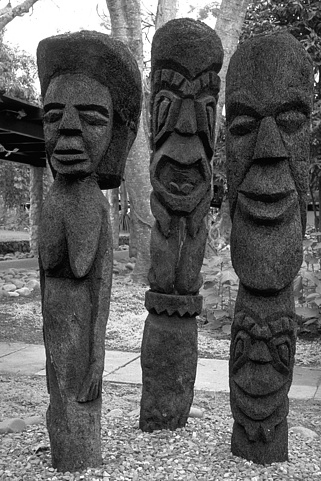}
}
\subfigure[Noisy image. Peak=2]{
\centering
\includegraphics[width=0.22\textwidth]{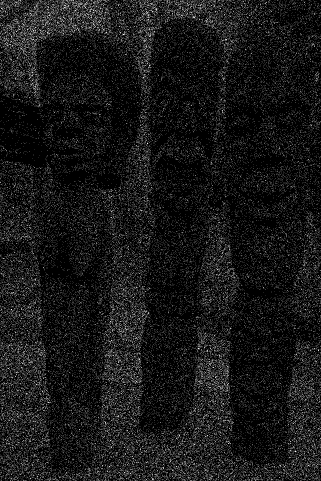}
}
\subfigure[NLSPCA (19.63/0.307)]{
\centering
\includegraphics[width=0.22\textwidth]{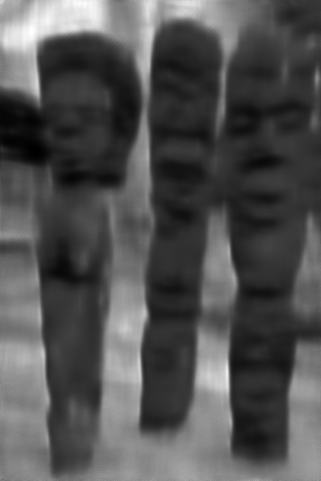}
}
\subfigure[NLSPCAbin (18.14/0.240)]{
\centering
\includegraphics[width=0.22\textwidth]{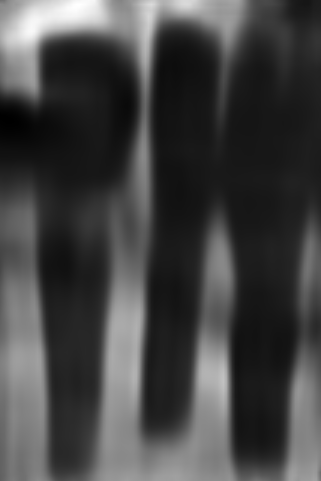}
}
\\
\subfigure[BM3D (20.16/0.371)]{
\centering
\includegraphics[width=0.22\textwidth]{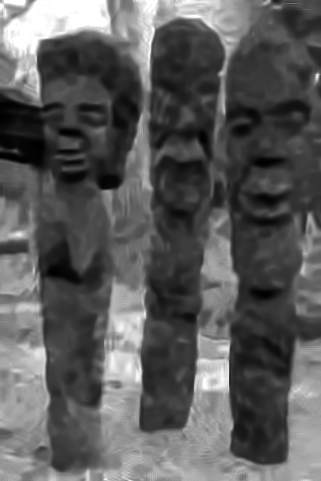}
}
\subfigure[BM3Dbin (19.95/0.340)]{
\centering
\includegraphics[width=0.22\textwidth]{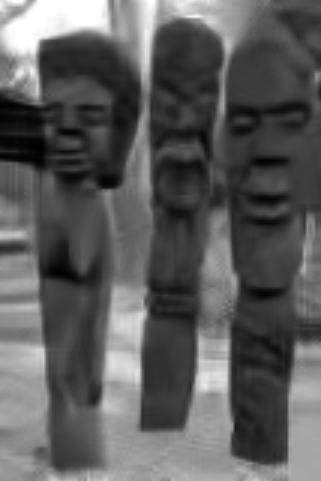}
}
\subfigure[$\mathrm{TRDPD}_{5 \times 5}^8$ (20.49/0.389)]{
\centering
\includegraphics[width=0.22\textwidth]{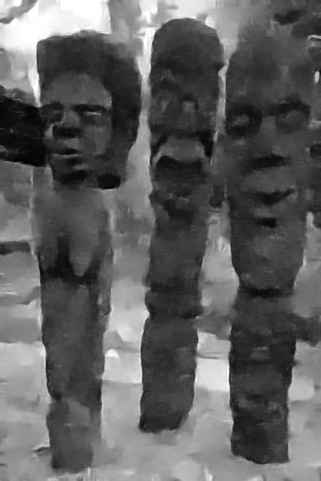}
}
\subfigure[$\mathrm{TRDPD}_{7 \times 7}^8$ (\textbf{20.52}/\textbf{0.393})]{
\centering
\includegraphics[width=0.22\textwidth]{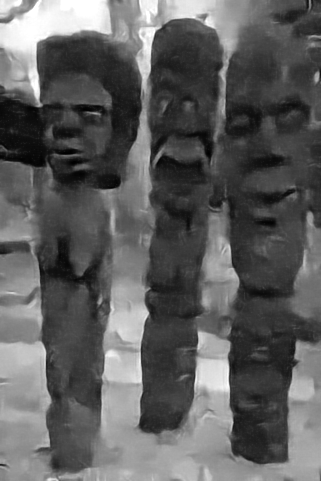}
}
\caption{Denoising of Image 3 with peak=2. The results are reported by PSNR/SSIM index. Best results are marked.}
\label{peak2}
\end{figure*}

\begin{figure*}[htbp]
\centering
\subfigure[Image 4]{
\centering
\includegraphics[width=0.22\textwidth]{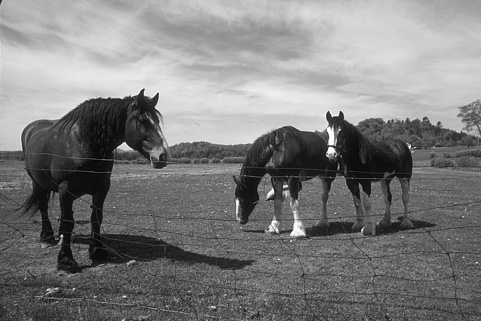}
}
\subfigure[Noisy image. Peak=2]{
\centering
\includegraphics[width=0.22\textwidth]{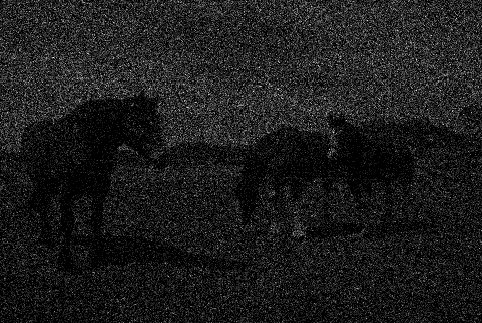}
}
\subfigure[NLSPCA (22.73/0.517)]{
\centering
\includegraphics[width=0.22\textwidth]{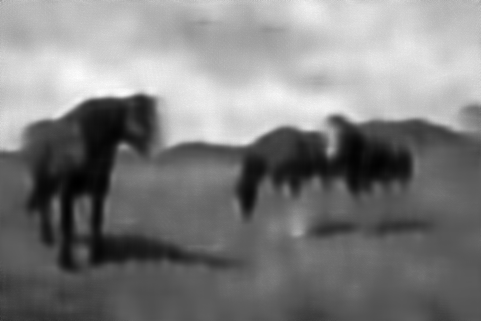}
}
\subfigure[NLSPCAbin (20.58/0.474)]{
\centering
\includegraphics[width=0.22\textwidth]{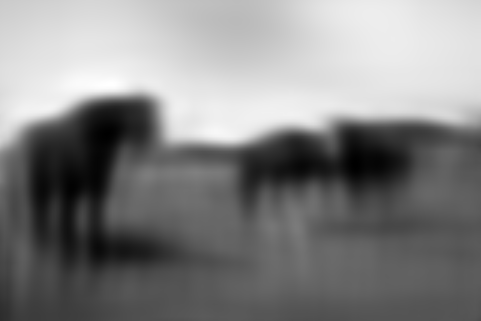}
}
\\
\subfigure[BM3D (23.10/0.502)]{
\centering
\includegraphics[width=0.22\textwidth]{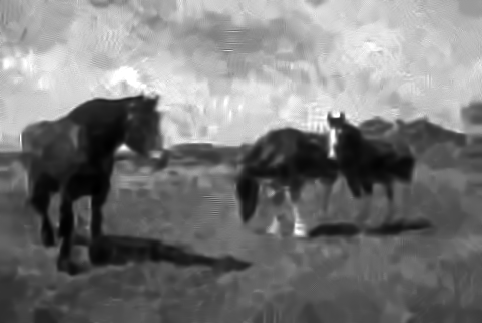}
}
\subfigure[BM3Dbin (23.24/0.537)]{
\centering
\includegraphics[width=0.22\textwidth]{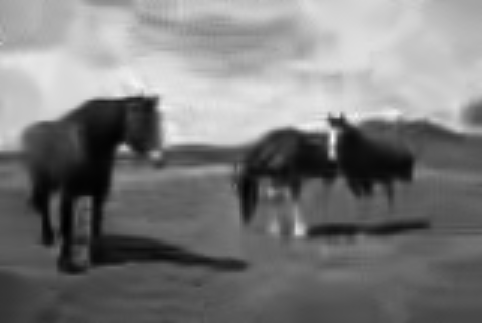}
}
\subfigure[$\mathrm{TRDPD}_{5 \times 5}^8$ (23.64/0.554)]{
\centering
\includegraphics[width=0.22\textwidth]{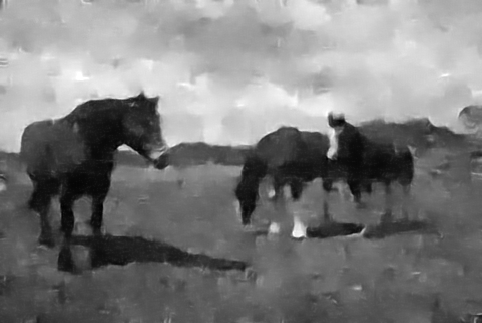}
}
\subfigure[$\mathrm{TRDPD}_{7 \times 7}^8$ (\textbf{23.74}/\textbf{0.558})]{
\centering
\includegraphics[width=0.22\textwidth]{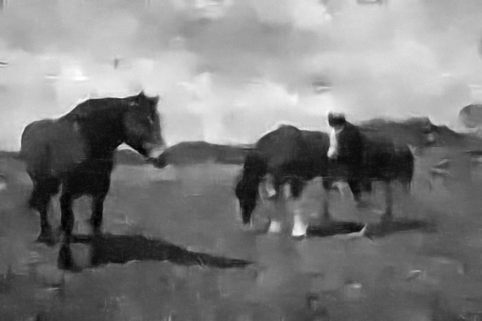}
}
\caption{Denoising of Image 4 with peak=2. The results are reported by PSNR/SSIM index. Best results are marked.}
\label{peak2-2}
\end{figure*}

\begin{figure*}[htbp]
\centering
\subfigure[Image 5]{
\centering
\includegraphics[width=0.22\textwidth]{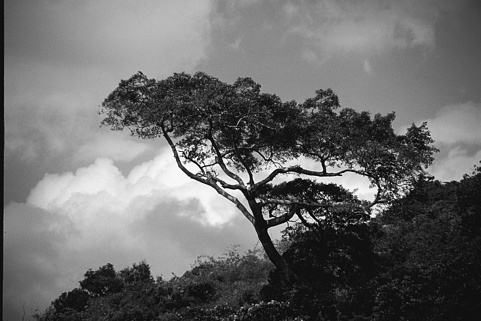}
}
\subfigure[Noisy image. Peak=4]{
\centering
\includegraphics[width=0.22\textwidth]{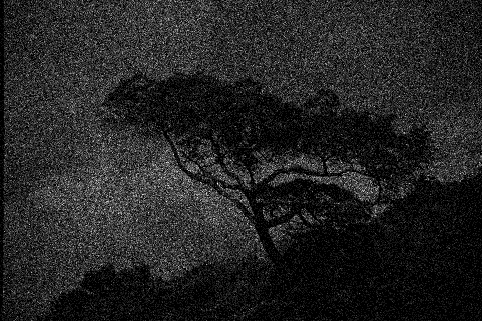}
}
\subfigure[NLSPCA (21.09/0.634)]{
\centering
\includegraphics[width=0.22\textwidth]{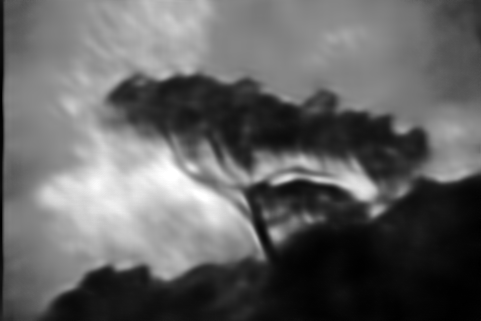}
}
\subfigure[NLSPCAbin (19.25/0.597)]{
\centering
\includegraphics[width=0.22\textwidth]{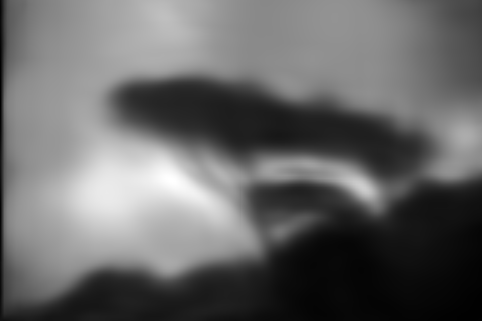}
}
\\
\subfigure[BM3D (23.29/0.655)]{
\centering
\includegraphics[width=0.22\textwidth]{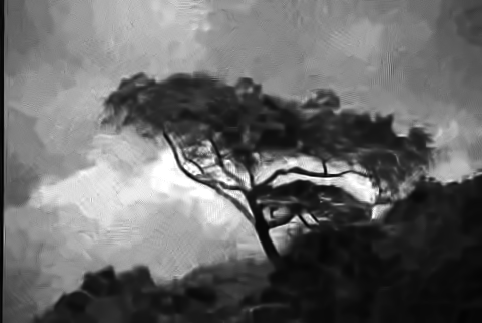}
}
\subfigure[BM3Dbin (22.22/0.666)]{
\centering
\includegraphics[width=0.22\textwidth]{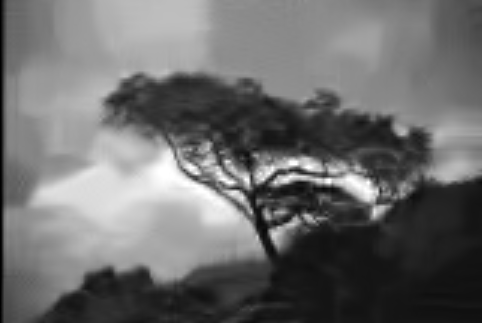}
}
\subfigure[$\mathrm{TRDPD}_{5 \times 5}^8$ (23.73/0.702)]{
\centering
\includegraphics[width=0.22\textwidth]{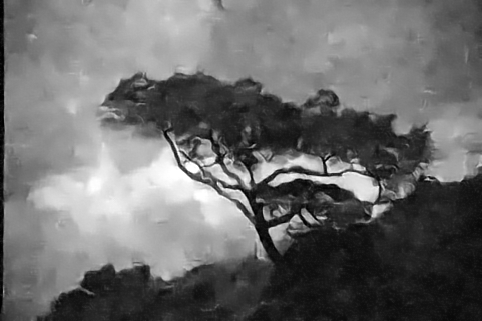}
}
\subfigure[$\mathrm{TRDPD}_{7 \times 7}^8$ (\textbf{23.83}/\textbf{0.714})]{
\centering
\includegraphics[width=0.22\textwidth]{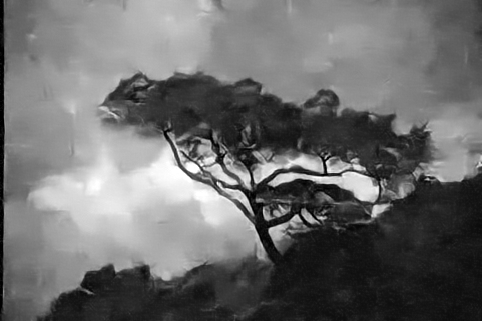}
}
\caption{Denoising of Image 5 with peak=4. The results are reported by PSNR/SSIM index. Best results are marked.}
\label{peak4}
\end{figure*}

\begin{figure*}[htbp]
\centering
\subfigure[Image 6]{
\centering
\includegraphics[width=0.22\textwidth]{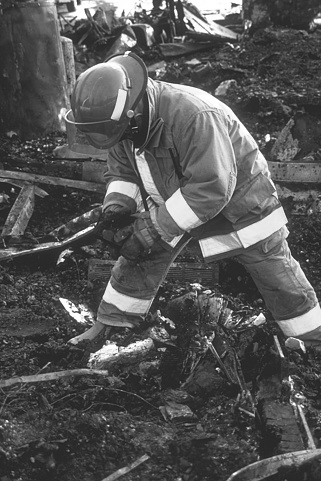}
}
\subfigure[Noisy image. Peak=8]{
\centering
\includegraphics[width=0.22\textwidth]{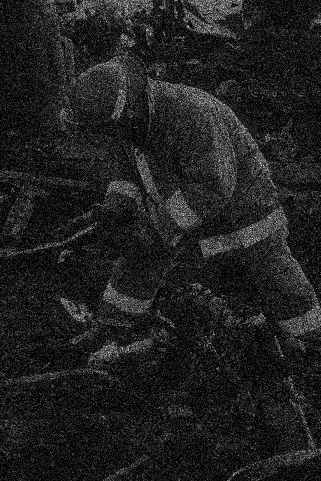}
}
\subfigure[NLSPCA (20.49/0.756)]{
\centering
\includegraphics[width=0.22\textwidth]{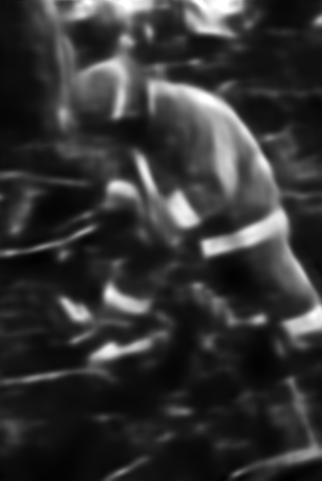}
}
\subfigure[NLSPCAbin (17.60/0.328)]{
\centering
\includegraphics[width=0.22\textwidth]{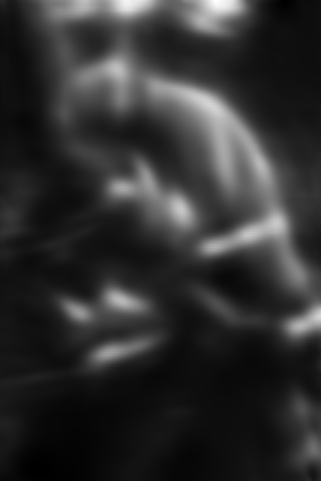}
}
\\
\subfigure[BM3D (23.20/0.596)]{
\centering
\includegraphics[width=0.22\textwidth]{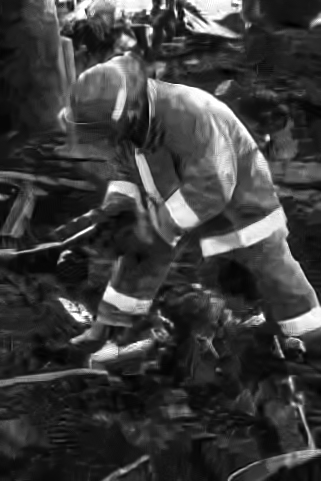}
}
\subfigure[BM3Dbin (21.62/0.499)]{
\centering
\includegraphics[width=0.22\textwidth]{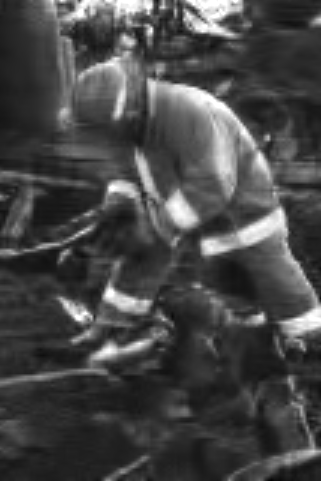}
}
\subfigure[$\mathrm{TRDPD}_{5 \times 5}^8$ (23.61/0.610)]{
\centering
\includegraphics[width=0.22\textwidth]{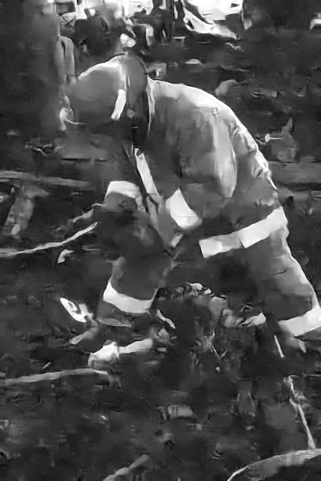}
}
\subfigure[$\mathrm{TRDPD}_{7 \times 7}^8$ (\textbf{23.68}/\textbf{0.617})]{
\centering
\includegraphics[width=0.22\textwidth]{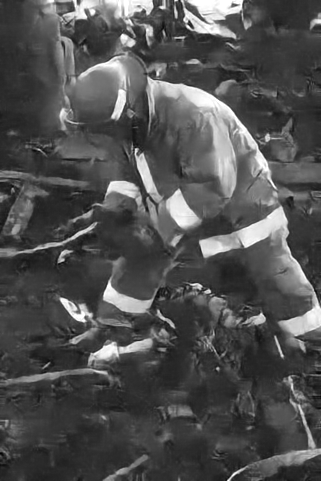}
}
\caption{Denoising of Image 6 with peak=8. The results are reported by PSNR/SSIM index. Best results are marked.}
\label{peak8}
\end{figure*}

In Fig.~\ref{peak2}-Fig.~\ref{peak4} are reported the recovered results for peak=2, 4 and 8 respectively. It can be observed that TRDPD and BM3D perform best on detail preservation, and achieve evidently better results in term of PSNR/SSIM index, with TRDPD even better. In the visual quality, the typical structured artifacts encountered with the BM3D-based algorithm do not appear when the proposed method TRDPD is used. Moreover, our method is more powerful in geometry-preserving, which can noticeably be visually perceived by comparison in Fig.~\ref{peak2}-Fig.~\ref{peak8}.

\begin{figure*}[htbp]
\centering
\subfigure[Image 7]{
\centering
\includegraphics[width=0.22\textwidth]{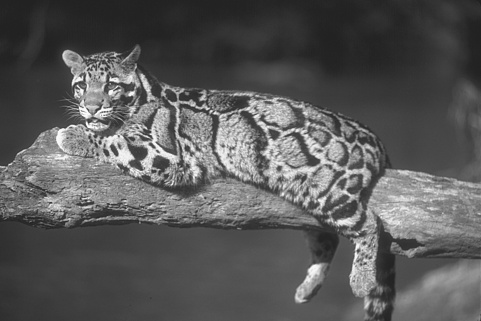}
}
\subfigure[Noisy image. Peak=20]{
\centering
\includegraphics[width=0.22\textwidth]{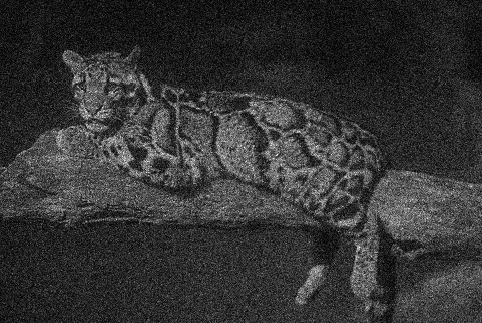}
}
\subfigure[NLSPCA (21.93/0.677)]{
\centering
\includegraphics[width=0.22\textwidth]{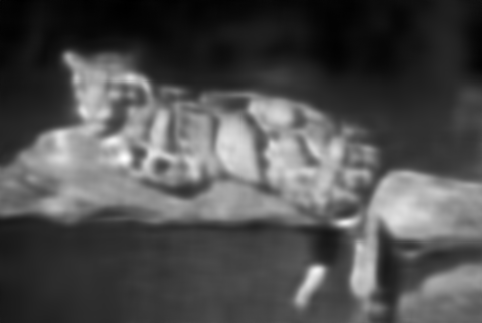}
}
\subfigure[NLSPCAbin (19.75/0.610)]{
\centering
\includegraphics[width=0.22\textwidth]{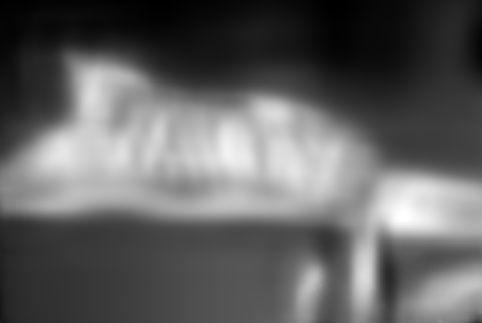}
}
\\
\subfigure[BM3D (26.76/0.847)]{
\centering
\includegraphics[width=0.22\textwidth]{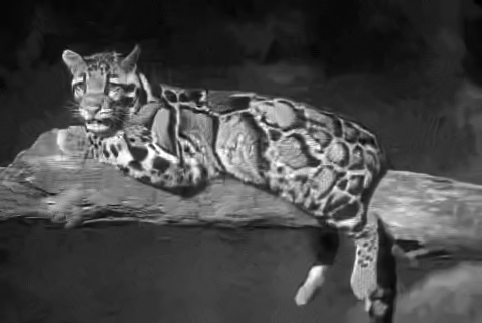}
}
\subfigure[BM3Dbin (24.33/0.775)]{
\centering
\includegraphics[width=0.22\textwidth]{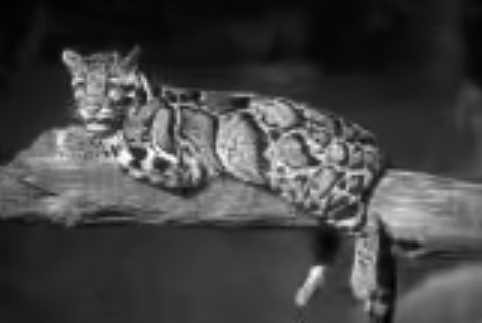}
}
\subfigure[$\mathrm{TRDPD}_{5 \times 5}^8$ (26.88/0.846)]{
\centering
\includegraphics[width=0.22\textwidth]{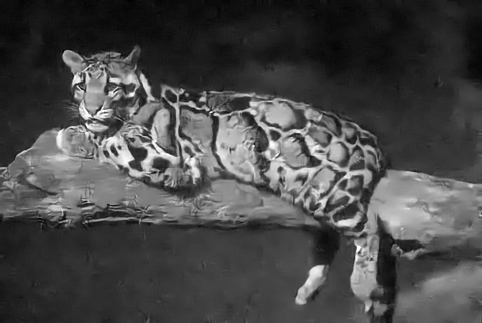}
}
\subfigure[$\mathrm{TRDPD}_{7 \times 7}^8$ (\textbf{27.01}/\textbf{0.856})]{
\centering
\includegraphics[width=0.22\textwidth]{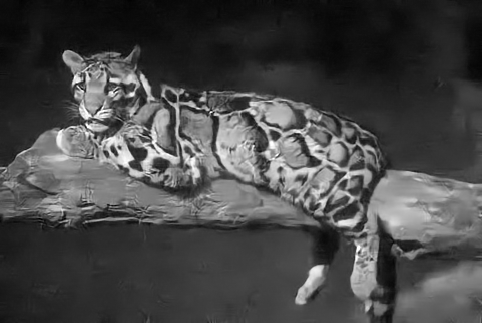}
}
\caption{Denoising of Image 7 with peak=20. The results are reported by PSNR/SSIM index. Best results are marked.}
\label{peak20}
\end{figure*}

\begin{figure*}[htbp]
\centering
\subfigure[Image 8]{
\centering
\includegraphics[width=0.22\textwidth]{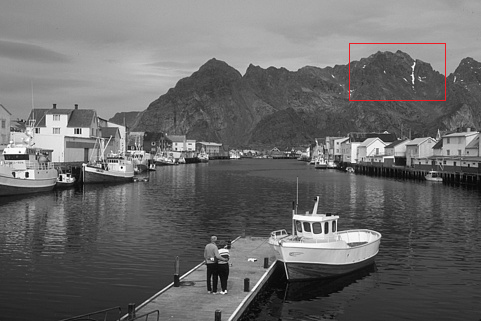}
}
\subfigure[Noisy image. Peak=40]{
\centering
\includegraphics[width=0.22\textwidth]{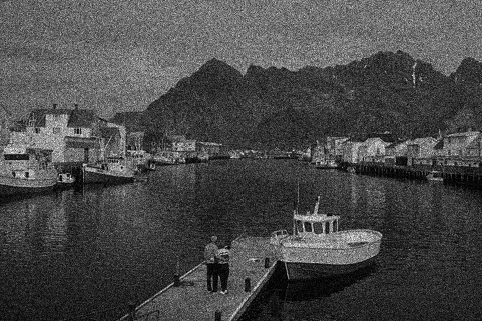}
}
\subfigure[NLSPCA (22.23/0.598)]{
\centering
\includegraphics[width=0.22\textwidth]{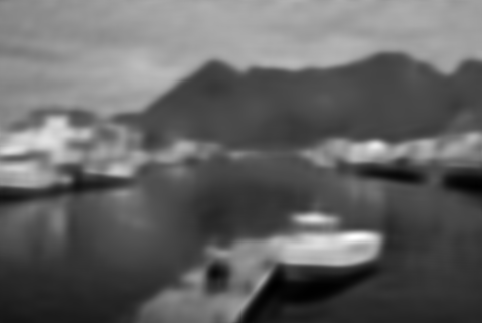}
}
\subfigure[NLSPCAbin (19.78/0.537)]{
\centering
\includegraphics[width=0.22\textwidth]{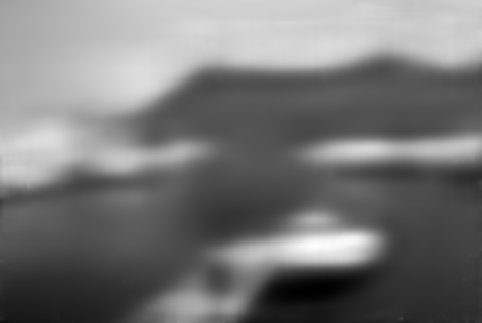}
}
\\
\subfigure[BM3D (28.67/0.799)]{
\centering
\includegraphics[width=0.22\textwidth]{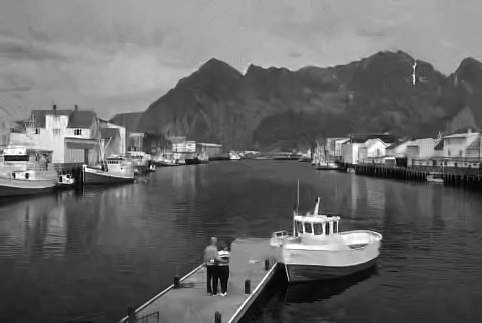}
}
\subfigure[BM3Dbin (24.25/0.678)]{
\centering
\includegraphics[width=0.22\textwidth]{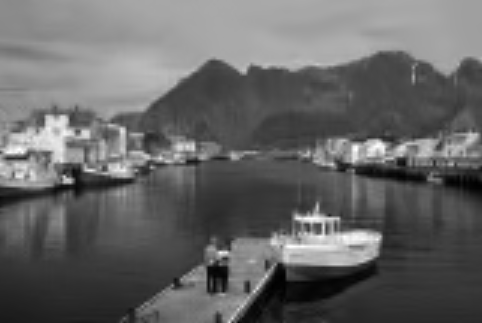}
}
\subfigure[$\mathrm{TRDPD}_{5 \times 5}^8$ (28.94/\textbf{0.807})]{
\centering
\includegraphics[width=0.22\textwidth]{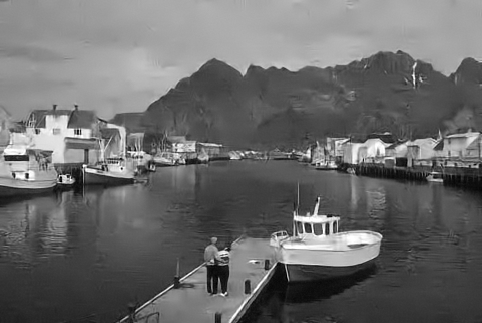}
}
\subfigure[$\mathrm{TRDPD}_{7 \times 7}^8$ (\textbf{29.06}/\textbf{0.809})]{
\centering
\includegraphics[width=0.22\textwidth]{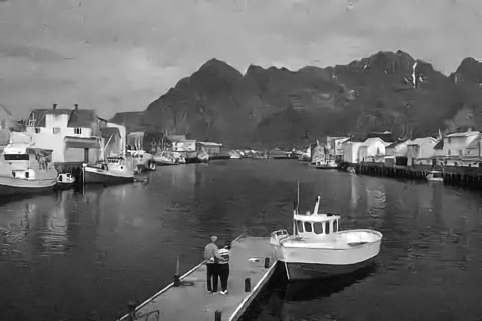}
}
\caption{Denoising of Image 8 with peak=40. The results are reported by PSNR/SSIM index. Best results are marked.}
\label{peak40}
\end{figure*}

We also presented the denoising results for relatively higher peak values, e.g., peak=20 and 40 in Fig.~\ref{peak20} and Fig.~\ref{peak40} respectively. By closely visual comparison, we can observe that TRDPD recovers clearer texture and sharper edges. The similar phenomenon can also be observed within the red rectangle shown in Fig.~\ref{peak40}(a), where TRDPD catches some tiny white features but BM3D-based method neglects them. Although these features are not quite obvious, the trained diffusion model still extracts them and exhibit these features apparently. In the TRDPD model, both the linear filters and influence functions are trained and optimized, whereby our model achieves some improvements over previous works. This critical factor of the optimized diffusion model is quite different from the FoE prior based variational model and traditional convolutional networks, where only linear filters are trained with fixed influence functions.

The recovery error in terms of PSNR (in dB) and SSIM are summarized in Table~\ref{resultshow}. Comparing the
indexes in Table~\ref{resultshow} and the denoising results in the present figures, the best overall performance is provided by the proposed method TRDPD. We also observe that $\mathrm{TRDPD}_{7 \times 7}^8$ can always gain an improvement of about (0.08$\sim$0.18dB)/(0.003$\sim$0.01) over $\mathrm{TRDPD}_{5 \times 5}^8$ in terms of PSNR/SSIM. From Table~\ref{resultshow}, one can see that the $\mathrm{TRDPD}_{7 \times 7}^8$ model outperforms the state-of-the-art BM3D-based method by (0.21$\sim$0.41dB)/(0.003$\sim$0.016).

\subsection{Run Time}
\begin{table}[t!]
\small
\begin{center}
\begin{tabular}{r|c|c|c|c}
\cline{1-5}
& BM3D & NLSPCA & $\mathrm{TRDPD}_{5 \times 5}^8$&$\mathrm{TRDPD}_{7 \times 7}^8$\\
\hline\hline
$256 \times 256$ & 1.38 & 367.9 &1.03 (\textbf{0.01})&2.43 (\textbf{0.02}) \\
$512 \times 512$ & 4.6 & 1122.1 &3.07 (\textbf{0.03})&7.45 (\textbf{0.07})\\
\cline{1-5}
\end{tabular}
\end{center}
\caption{Typical run time (in second) of the Poisson denoising methods for images with two different dimensions.
The CPU computation time is evaluated on Intel CPU X5675, 3.07GHz.
The highlighted number is the run time of GPU implementation based on NVIDIA Geforce GTX 780Ti.}
\label{runtime}
\end{table}

It is worthwhile to note that our model merely contains
convolution of linear filters with an image, which offers high levels of parallelism
making it well suited for GPU implementation.

In Table \ref{runtime}, we report the typical run time of our model
for the images of two different dimensions for the case of $\text{peak} = 4$.
We also present the run time of two competing algorithms for a comparison \footnote{
All the methods are run in Matlab with single-threaded computation for CPU implementation.
We only consider the version without binning technique.}.

Due to the structural simplicity of our model, it is well-suited to GPU parallel computation. We are able to implement our algorithm on
GPU with ease. It turns out that the GPU implementation
based on NVIDIA Geforce GTX 780Ti can accelerate the inference procedure significantly, as shown in Table \ref{runtime}. By comparison, we see that our TRDPD model is generally faster than the other methods, especially with GPU implementation.

\section{Conclusion}
In our study we exploited the newly-developed trainable nonlinear reaction diffusion model in the context of Poisson noise reduction.
Its critical point lies in the both training of filters and the influence functions in the reaction diffusion model by taking into account the Poisson noise statistics. Based on standard test dataset, the proposed nonlinear diffusion model provides strongly competitive results against state-of-the-art approaches, thanks to its several desired properties: anisotropy, higher order and adaptive forward/backward diffusion through the learned nonlinear functions. Moreover, the proposed model bears the properties of simple structure and high efficiency, therefore is well suited to GPU computing.

In our current work, the trained diffusion process is targeted for natural images.
However, the Poisson noise often arises in applications such as
astronomy imaging, biomedical imaging and fluorescence microscopy.
Therefore, training specialized diffusion process for specific images bears the potential to
improve the current results. This could be our future study.
{
\small
\bibliographystyle{ieee}
\bibliography{references}
}

\end{document}